\pdfoutput=1
\documentclass[11pt]{article}

\usepackage[final]{acl}

\usepackage{times}
\usepackage{latexsym}
\usepackage{tikz}

\usepackage[T1]{fontenc}
\usepackage[utf8]{inputenc}

\usepackage{microtype}

\usepackage{inconsolata}

\usepackage{graphicx}
\usepackage{xspace}
\usepackage{xcolor}
\usepackage{comment}
\usepackage[inline,shortlabels]{enumitem}

\usepackage{subcaption}
\usepackage{booktabs}
\usepackage{rotating}

\newcommand{\invisible}[1]{}

\newcommand{\dataname}{VISaGE\xspace}

\newcommand{\genimg}{generic\xspace}
\newcommand{\exptimg}{exception\xspace}
\newcommand{\expttxt}{$e_{c,a}$\xspace}
\newcommand{\gentxt}{$n_{c,a}$\xspace}
\newcommand{\genimgvar}{$V_c$\xspace}
\newcommand{\exptimgvar}{$V_e$\xspace}

\newcommand{\generic}[1]{\textit{#1}}
\newcommand{\category}[1]{\texttt{#1}}
\newcommand{\model}[1]{\texttt{#1}}

\newcommand{\circone}{\raisebox{.5pt}{\textcircled{\raisebox{-.9pt} {1}}}\xspace}
\newcommand{\circtwo}{\raisebox{.5pt}{\textcircled{\raisebox{-.9pt} {2}}}\xspace}

\title{\dataname: Understanding Visual Generics and Exceptions}

\author{Stella Frank \\
  University of Copenhagen, DK \\
  \texttt{stfr@di.ku.dk} \\\And
  Emily Allaway \\
  University of Edinburgh, UK \\
  \texttt{emily.allaway@ed.ac.uk} \\}

\begin{document}
\maketitle
\begin{abstract}
While Vision Language Models (VLMs) learn conceptual representations, in the form of generalized knowledge, during training,
they are typically used to analyze individual instances. 
When evaluation instances are atypical, this paradigm results in tension between two priors in the model. The first is a \textit{pragmatic prior} that the textual and visual input are both relevant, arising from VLM finetuning on congruent inputs; the second is a \textit{semantic prior} that the conceptual representation is generally true for instances of the category.
In order to understand how VLMs trade off these priors, we introduce a new evaluation dataset, \dataname, consisting of both typical and \textit{exceptional} images.
In carefully balanced experiments, we show that
conceptual understanding degrades
when the assumption of congruency underlying the pragmatic prior is violated with incongruent images.
This effect is stronger than the effect of the semantic prior when querying about individual instances.
\end{abstract}

\section{Introduction}

Vision-language models (VLMs) are typically used to analyze \emph{instances}: what is going on in a particular image?
Moreover, during training they learn a set of \emph{conceptual} representations: generalized knowledge that holds over many instances.
However, \emph{exceptions} to generalizations, in the form of atypical instances, disrupt the alignment of in-context instance understanding and in-weights conceptual knowledge.
While VLMs have been thoroughly tested on their ability to discern minimal differences between image instances~\citep[e.g.,][]{johnson2017,thrush2022,tong2024c},
and likewise their conceptual representations (based on exposure to typical instances) have also been analyzed~\cite[e.g.,][]{bruni2014,silberer2013,collell2016,oneata2025},
the potential tension between instance and concept representations, as arises in atypical instances, is currently under-explored.

In language, the attributes associated with a conceptual category are often expressed through \textit{generics}: generalizations without quantifiers (e.g.,~\generic{cats have four legs}). 
This lack of quantification means that generics remain true regardless of exceptions (tripod cats---cats missing one leg---do not impact the truth of \generic{cats have four legs}).
In other words, the attribute is associated as characteristic of the category regardless of how frequent it actually is.\footnote{
This is a substantial simplification of the semantics of generics~\citep[cf.][]{krifka1987outline}.}

\begin{figure}[t]
\includegraphics[width=\columnwidth]{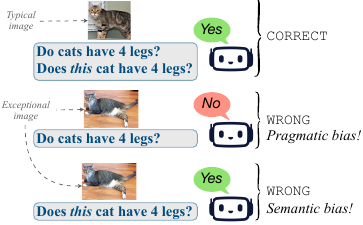}
  \caption{\dataname contains both typical and exceptional instances of a category, with respect to some conceptual norm. 
  % for which generics do not hold.
  We probe VLMs for conceptual and instance-level understanding,
  with congruent (top) and incongruent (middle, bottom) text-image pairs. Model failures indicate effects of both pragmatic and semantic biases.
  % which is congruent in the typical case (top) but
  % conflicts in the case of exceptional instances of a category~\category{cat} (middle).
  % However, the same exceptional instance can also be a typical member of the exception category (bottom pair).
  }
  \label{fig:fig1}
\end{figure}

Unlike language, which can both denote on this generic or conceptual level, as well as refer to a particular instance, VLMs are always grounded in a particular visual instance.
Work that has probed for conceptual attributes %~\cite{bruni2014,collell2016}
has used typical instances to stand in for the concept.
This conflates instance and conceptual representations.
In order to separate the two, visual \emph{exceptions} are required: instances of a category that violate the generic (see Figure~1).

To this end, we introduce a new evaluation dataset, \textbf{VisaGE:} Visual Generics and Exceptions, consisting of 
conceptual categories with images of both typical and exceptional instances.
Specifically, exceptions are always with regard to a particular generic norm,~i.e., a typical attribute: a \category{tripod cat} is an exception for \generic{cats have four legs}, but is typical for \generic{cats have a long tail}.
The category-attribute pairs in \dataname, along with their exceptions, are extracted from textual generics and carefully manually validated, together with the image instances.

Using \dataname, we investigate two questions:
\begin{enumerate}
\setlength{\itemsep}{0em}
\item \textbf{(RQ1)} How does visual grounding to (potentially atypical) instances impact a model's ability to access \emph{conceptual information}?
\item \textbf{(RQ2)} How does conceptual information, as recruited by text labels, impact VLMs' ability to recognize \emph{instance attributes}? 
\end{enumerate}
These research questions examine the effects of two priors in VLMs. 
The first is a \textit{pragmatic prior}, arising from VLM finetuning, that the textual and visual input are congruent and both relevant; the second is a \textit{semantic prior} that the category-attribute generic is generally true.\footnote{This is analogous to the Gricean maxims of relevance and quality (truthfulness)~\citep{grice1975logic}.}
In the exceptional image settings we explore with \dataname, these two priors can conflict:
when asking about conceptual knowledge (RQ1),
the atypical image must be ignored and the semantic prior followed,
while for instance queries (RQ2),
given an atypical instance, the pragmatic prior to focus on the current context must overrule the semantic prior of typicality.

We test a set of contemporary open-weight VLMs and find evidence that their conceptual representations do not recognize possible variation in attributes.
We observe that the models' pragmatic prior interferes with conceptual understanding (and the semantic prior) when visual grounding is incongruent with the text.
This suggests that VLMs do not correctly differentiate between generic concepts and specific instances.
Results are more mixed when models are tasked to recognize instantiations of exceptional attributes in images: while most models do show evidence of a semantic bias, this effect is less strong.

Our contributions are:
\begin{enumerate*}[(1)]
\item a new dataset, \dataname, consisting of concept-attribute pairs with images of 
both typical (generic) and exceptional instances;
\item experimental evidence that VLM conceptual representations are visually grounded only in typical or generic instances and are not sufficiently robust to within-category variation. 
\end{enumerate*}

\section{Background}
Previous work has investigated the semantics of generics with LMs~\citep{ralethe2022,collacciani-etal-2024-quantifying,cilleruelo-etal-2025-generics}. These studies show LMs often struggle to account for and reason about exceptions in both probing~\citep{allaway2024} and reasoning~\citep{allaway-mckeown-2025-evaluating} tasks. However, they have not considered 
generics in VLMs, particularly how visual grounding interacts with generic's semantics.

For evaluating VLMs, most visual benchmarks test situational and configurational instance understanding~\citep{thrush2022,li2024c}, sometimes with (synthetic) atypical examples~\cite{bitton-guetta2023}. Although
\citet{saleh2013} create a small dataset
of exceptional object images, these are not annotated with semantic attributes, unlike \dataname.
More recently,~\citet{luo2025,vo2025} create datasets using synthetic image generation to manipulate object attributes.

Additionally, our experiments, in which we manipulate image--text congruency,
contribute to a line of work investigating the relative importance of different modalities in VLMs~\cite{gat2021,frank2021, fu2024a,tong2024c,li2024c,parcalabescu2025}.

\section{Dataset}
\label{sec:dataset}

Our dataset \dataname is constructed 
by first collecting text pairs (\gentxt, \expttxt) where \gentxt is a conceptual norm for category $c$ with attribute $a$ and \expttxt is an exception to that norm (i.e., a subcategory of $c$ that does not have the attribute $a$).
Then for each pair, we retrieve two sets of images corresponding to cases where the norm applies (\genimg images \genimgvar) and where it does not (\exptimg images \exptimgvar). The resulting dataset then consists of tuples (\gentxt, \expttxt, \genimgvar, \exptimgvar).
Finally, we manually validate and expand the dataset (details in Appendix~\ref{sec:appsecdataset}).

\dataname contains 1601 exceptional image examples for 437 exception subcategories, derived from 296 category-attribute relations (generics/conceptual norms) for 171 categories,
balanced with the same number of typical images.\footnote{
Dataset and code at \url{github.com/scfrank/visage1601}}

\begin{figure*}[!t]
  \includegraphics[width=0.9\textwidth]{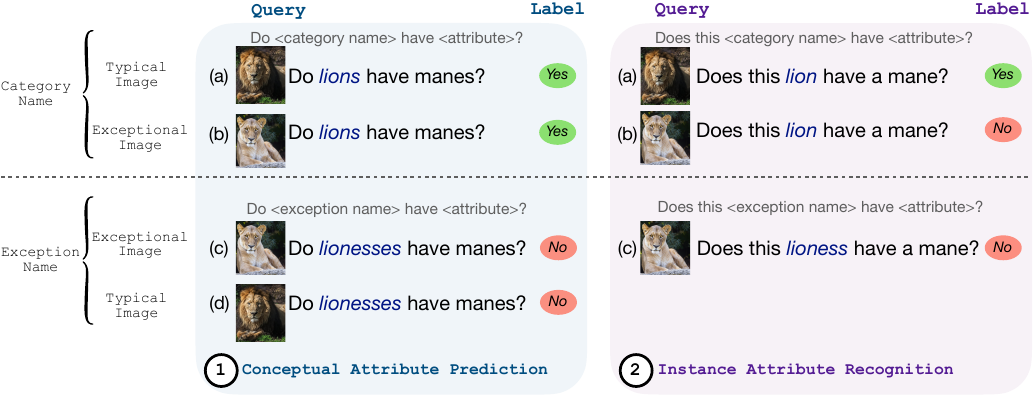}
  \vspace{-0.5em}
  \caption {Summary of experiments and conditions: 
  Exp.~\circone tests models' conceptual understanding of generics, while Exp.~\circtwo tests models' ability to reason about instances.
  }
  \label{fig:experiments}
\end{figure*}

\paragraph{Norm-Exception Text Pairs}

For our initial set of concept-attribute norms, we intersect the category-attribute lists of XCSLB~\citep{devereux2014,misra2022property} and the McRae norms~\citep{mcrae2005}, with the categories in the THINGS object image dataset~\cite{hebart2019}.
This results in a robust set of conceptual norms expressed as generics.
Then, for each generic (category-attribute statement) we generate a set of exceptions \expttxt using the LM prompting framework proposed by~\citet{allaway2024}.
We retain the short exceptions, ideally corresponding to subcategories.

\paragraph{Images}
We retrieve a large set of images for each exception subcategory using Bing Image Search by querying for the exception name~\expttxt.
Subsequent human validation (see below) selects the best images, resulting in a mode of 4 images per exception.
A matched number of generic images for each category are taken from the THINGS dataset. %~\citep{hebart2019}.
THINGS images were specifically collected to be typical object instances; we further manually validate the applicability of the generic conceptual norms.

\paragraph{Validation}
We collect three types of validation annotations for each tuple. % (\gentxt, \expttxt, \genimgvar, \exptimgvar).
First, we validate that the images \genimgvar retrieved from THINGS exhibit the conceptual norms \gentxt.
Second, we validate that each \expttxt is actually an exception to the norm \gentxt. With this we filter out exception subcategories that are hallucinated (e.g., \category{strawberry blonde cheetah}) or incorrectly related (e.g., not exceptional or not actually subcategories).
Finally, we validate that the retrieved images \exptimgvar for each exception are correct. We exclude images that are the wrong category (e.g.,~images of Ryan Gosling retrieved for the category \category{gosling}) or that are the wrong style
(e.g.,~not object-centered photographs).

\section{Experiments}
\label{sec:experiments}

Using \dataname,
our experiments query VLMs about conceptual and instance attributes across a number of conditions: see Figure~\ref{fig:experiments} for an overview.
Specifically, we vary
\begin{enumerate*}[(1)]
\item the type of knowledge being queried (conceptual vs.~instance);
\item the type of image input (typical vs~exceptional images); and
\item the noun-phrase used to refer to the concept (category-name vs~exception-name reference).
\end{enumerate*}

\begin{figure*}[ht]
\includegraphics[width=\textwidth]{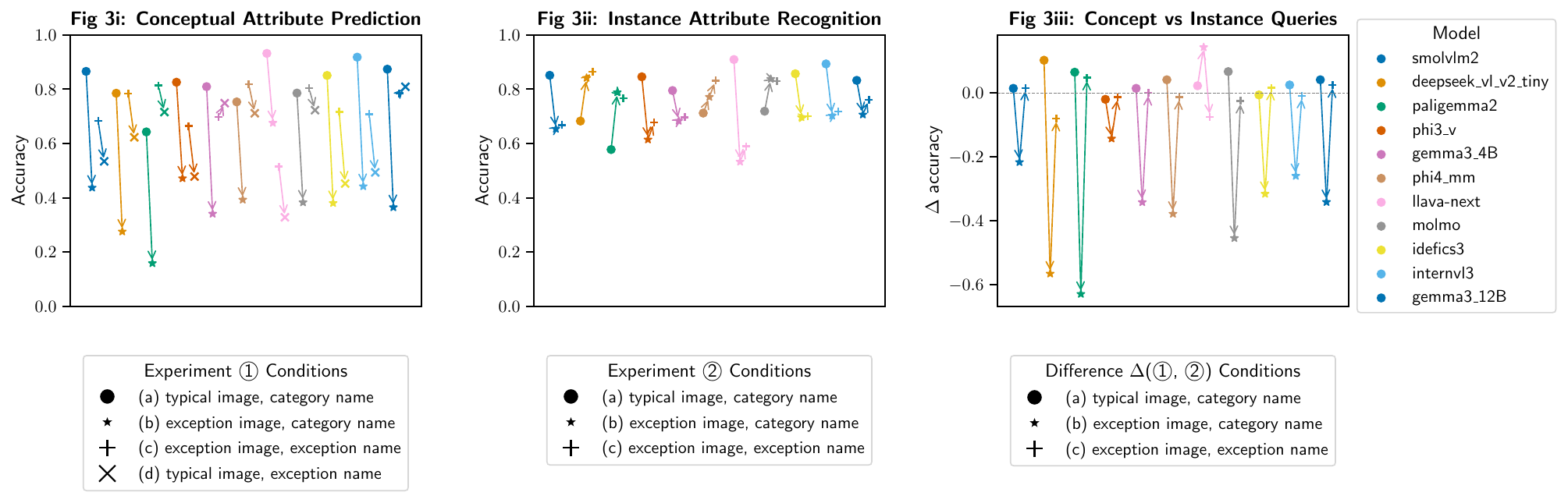}
     \caption{Results: See Fig.~\ref{fig:experiments} for setup.
     Exp~1:~Conceptual attribute prediction accuracy decreases for incongruent inputs~(b and d).
     Exp~2:~Across most models, instance attribute recognition declines for exceptional images~(b), unless they are named as such~(c).
     Fig 3iii shows data from Exps 1\&2:~The difference between conceptual and instance accuracy is highest for incongruent pairs~(b).
     Most models have higher instance accuracy in condition~(b).
     Models are ordered by size.
     Numbers in Appendix~\ref{app:full-results}.
     }
     \label{fig:all_results}
\end{figure*}

\paragraph{Models}
We test a suite of open-weights VLMs:
these are listed in Appendix~\ref{app:models}.
We use the \texttt{vllm} library to wrap our prompts\footnote{Conceptual prompt template example: \texttt{Answer yes or no. Do \{concept-pl\} have \{attribute\}?}\\ Instance prompt template example: \texttt{Answer yes or no. Does this \{concept-sg\} have \{attribute\}?}} in the correct model-specific formats.

\paragraph{Evaluation}
We report the percentage of correct (yes/no) responses for each model, using the first token of the model output.
Note that the correct response depends on the condition: see~Figure~\ref{fig:experiments}.

\subsection{Conceptual Attribute Prediction}
\label{subsec:exp2}
Our first experiment \circone targets RQ1 and examines how visual groundings impacts model responses to queries about conceptual information.  
Specifically, we use two pairs of conditions to investigate the impact of text-image congruency (see Fig.~\ref{fig:experiments}~\circone).
The first pair of conditions uses the category name in conceptual queries (``Do lions have manes?''): (1a) typical (congruent), and (1b) exceptional (incongruent) images. The second pair of conditions similarly queries conceptual information but about the \textit{exception} subcategory (``Do lionesses have manes?''): (1c) exceptional (congruent), and (1d) typical (incongruent) images. 

Our results (Fig.~\ref{fig:all_results}i) show that VLMs' ability to answer conceptual questions degrades when the visual grounding is incongruent with the text input. That is, we observe a drop in accuracy 
from congruent to incongruent inputs in both pairs of conditions
((1a) $\rightarrow$ (1b) and (1c) $\rightarrow$ (1d)).
This suggests that the \textit{pragmatic prior}, i.e.,~assuming the image is relevant, is overriding the correct retrieval of conceptual knowledge.

Intriguingly, we also observe that
incongruency in the input has less impact (interferes less) on accuracy when the queries are about the exception subcategory, and the incongruent image is of a typical instance, rather than vice-versa.
This could be due to models' processing atypical images more attentively than typical images; we return to this in Section~\ref{sec:shapley}.

\subsection{Instance Attribute Recognition}
\label{subsec:exp1}
Our second experiment \circtwo aims to answer RQ2 by examining how conceptual information impacts VLMs' ability to answer instance-specific queries.
We specifically focus on the role of language-based conceptual activation; that is, can VLMs override conceptual generalizations to recognize specific attributes of individual instances. 
We compare category-name instance queries (``Does this lion have a mane?'') in two conditions: with (2a) typical and (2b) exceptional images (see Fig.~\ref{fig:experiments}~\circtwo). 
The third condition (2c) uses the \textit{exception-name} in instance queries
with exceptional images,
providing an explicit language cue to the model to consider the exception rather than the category conceptual representation.

Our results (Figure~\ref{fig:all_results}ii) indicate that,
despite instance queries directing the model to consider the image, 
many
models still appear to ignore the visual features, relying instead on language-based conceptual cues.
Specifically,
we observe a drop in accuracy from condition (2a) to (2b).
When the text and image are \textit{in}congruent (in (2b)), the conceptual information activated by the category name (semantic prior) does not apply to the image, since the image is exceptional.
Inasmuch as a model is relying on its semantic prior, this will result in a substantial drop in accuracy.
Note that if the models instead prioritized using the visual features of the input, performance would be relatively stable across conditions.

Congruence alone does not explain the results:
Despite the image and text being congruent in (2c), the accuracy is comparable to (2b),
indicating that presenting the correct semantic information is of limited use for improving instance attribute recognition in atypical instances.
One reason for this could be that the exception categories are lower frequency than the general categories, so exception attribute knowledge may be less well developed (see also (1a) vs.~(1c)). 
In addition, many of the exception names include the category name (e.g.,~\textit{lion}ess, tripod \textit{cat}), which could lead to semantic priming of the general category and its attributes.

Not all models show semantic bias effects: some are more accurate when queried for exception images, rather than typical images.
We speculate that these VLMs are sensitive to the (a)typicality of images and are also considering the pragmatics of the query.
That is, since the query is asking about the application of a conceptual norm to a specific instance, there must be something special about that instance: it might be expected to be a violation of the norm.
This behavior would lead to lower accuracy in (2a), where the image isn't exceptional, and higher accuracy in both (2b) and (2c), which we observe with four of the VLMs we evaluated.

Finally, we observe (Figure~\ref{fig:all_results}iii) that the effect of pragmatic prior violation is stronger than violations of the semantic prior.
When visual input and category name are congruent ((a) and (c)), models perform similarly for both conceptual and instance queries (difference is near-zero).
In contrast, when models are required to disregard the image (conceptual queries), performance suffers significantly, compared to instance queries in which the semantic prior is incongruent with the exception image (condition (b): difference is negative).
The accuracy difference that is visible \textit{only} with incongruent inputs
emphasizes the importance of considering exceptional examples to test how image instances interact with conceptual representations.

\subsection{Feature Attribution with Shapley Values}
\label{sec:shapley}

\begin{figure}[t]
  \includegraphics[width=\linewidth]{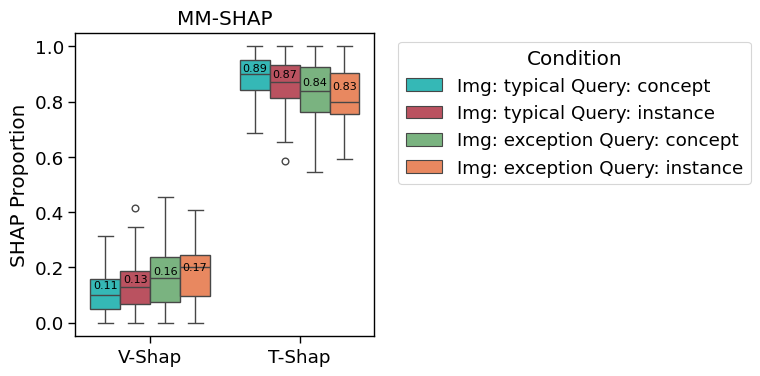}
  \caption {MM-SHAP calculated with \model{smolVLM2}.
  V-SHAP measures the proportion of Shapley values coming from the image, while T-SHAP is the proportion from text tokens.}
  \label{fig:shapely:smolvlm}
\end{figure}

We perform a Shapely value analysis \cite{shapley1952,lundberg2017} in order to understand how different parts of the input contribute to model predictions.
Specifically, does the image contribute more to the model prediction in the instance query condition, compared to the concept query condition?

In our setting, the Shapley value represents the contribution of a single input feature (e.g.,~a text token or the image, which we process as a single feature) to the model's prediction using the full input (i.e.,~the generation of a ``yes''  or ``no'' token in response to the query).
Following MM-SHAP~\citep{parcalabescu2023,parcalabescu2025}, we calculate the proportional contribution of the image input (V-SHAP) and the text tokens (T-SHAP) aggregated over the dataset.
See~Appendix~\ref{app:shapley} for details.

We calculate MM-SHAP values for a small model (\model{smolvlm2}) that behaved similarly to other models in our experiments, comparing instance and conceptual queries with typical and exception images.
The results (Figure~\ref{fig:shapely:smolvlm}) do not show a qualitative difference in use of the image between concept and instance conditions, confirming a general pragmatic bias of attending to the image in all contexts.
However, quantitatively, there is an increase in V-SHAP for each instance query condition, compared to the corresponding concept query condition (all differences are significant at $p<0.01$ with a paired t-test; effect size, as measured with Cohen's $d$, between the two typical image conditions is $d=0.3$, while for the exception image conditions $d=0.19$, both indicating a small effect).
Furthermore, conditions with exception images have higher V-SHAP compared to typical images
($p<0.01$; $d=0.62$ for concept queries and $d=0.56$ for instance queries, both indicating a medium effect),
suggesting that the models are recruiting visual information to supplement and possibly counteract the conceptual information from the text, in the cases where these are incongruent or atypical.

\section{Conclusion}
VLMs must balance learned priors with the requirements of the current context.
With the use of a new dataset of visual exceptions, \dataname, 
we have shown that VLMs have not yet solved this task: models neither reliably attend to the exception instance, ignoring the conceptual semantic prior, nor can they reliably ignore distractor images to answer generic conceptual queries.

\clearpage

\section*{Limitations}

Our categories and attributes are limited to conceptual norms in American English. This is because the typical images we use for visual grounding (derived from THINGS) are based on American English definitions of categories.
Conceptual spaces are language-dependent and different languages will make different conceptual distinctions, attending to different attributes.
However, we believe the general patterns of results would hold across languages and models, since the distinction between instance-level and conceptual-level reasoning is common across languages.

The data collection process focused on quality rather than recall; we may have inadvertently omitted particular important exception types. 
In particular, exceptions that are rare, hard to see, or unlikely to be photographed, are missing (e.g.,~\category{insomniac owl} as an exception for \generic{owls sleep in the day}, \category{cheetah with a broken leg} as an exception for \generic{cheetahs are fast}).

\paragraph{Risks} The concepts in our dataset correspond to concrete object categories.
However, the difficulty of appropriately distinguishing (exceptional) instances vs.~conceptual generalizations can also apply to categories that group people, where overgeneralization can lead to stereotyping.
Understanding VLM capabilities and limitations is a step towards mitigating these risks.

\section*{Acknowledgments}
This work was supported in part by the Pioneer Centre for AI, DNRF grant number P1 and an Edinburgh–Copenhagen Strategic Partnership Award.

\bibliography{visualgenerics,stellavisualgenerics}

\appendix

\section{Dataset Construction}
\label{sec:appsecdataset}
The McRae norms are conceptual norms elicited from humans~\cite{mcrae2005}. \citet{devereux2014} builds on these in the XCSLB dataset and then \cite{misra2022property} further revise them. Each norm can be expressed as a generic.

To generate exceptions to the conceptual norms, we use the framework proposed by \citet{allaway2024}. This framework proposes specific prompt templates for generating exceptions from LLMs, along with a filtering process to ensure the generated exceptions are true and salient. We use these templates with GPT-3.5~\cite{ouyang2022training}\footnote{\texttt{{gpt-3.5-turbo-0613}}} to generate candidate exceptions and remove false ones. We keep the top 5 candidates ranked by perplexity to use in our dataset.  

\dataname includes substantial human validation, including an iterative process of adding new attribute norms and exceptions.
During validation, annotators can revise and expand the dataset by adding additional exceptions and category-attribute relations. Specifically, for valid category-attribute relations annotators, can provide an additional exceptional subcategory $\hat{e}_{c,a}$. Additionally, for each exception, annotators can provide a new category-attribute relation $n_{c,\hat{a}}$ that the exception corresponds to. This allows us to capture subcategories that are exceptional for the category but not for the original attribute $a$. For example, pixie-bob cats are an exception to \generic{cats have long tails} but not to the original norm \generic{cats have tails}. The tuples with the new category-attribute norms ($n_{c,\hat{a}}$, $e_{c,\hat{a}}$, \genimgvar, \exptimgvar)\footnote{Note that $e_{c,\hat{a}} = e_{c,a}$; the changed index is for clarity.} are added directly into the dataset while for the new exceptions $\hat{e}_{c,a}$, new images $V_{\hat{e}}$ are first retrieved and validated before being added to the dataset as (\gentxt, $\hat{e}_{c,a}$, \genimgvar, $V_{\hat{e}}$). 

The annotations were conducted by the authors of this paper. Through the revision and expansion process, we added 121 new tuples of conceptual-norm-and-exception (along with their corresponding images). Combined with the added conceptual norms, we nearly doubled the size of our dataset (an increase from 872 tuples to the final 1601 tuples).

\section{Models}
\label{app:models}

\begin{table*}[t]
\footnotesize
  \begin{tabular}{llr}
    \toprule
    Short Name & HF Model Name & Size \\ % &  Visual Encoder & LLM \\
    \midrule
    \model{smolvlm2}   &  \model{HuggingFaceTB/SmolVLM2-2.2B-Instruct} & 2.2B  \\%& google/siglip-so400m-patch14-384 & HuggingFaceTB/SmolLM2-1.7B-Instruct \\
    \model{deepseek-vl-v2} &  \model{deepseek/deepseek-vl2-tiny}  & 3.37B    \\%& SigLIP-SO400M-384 + tiling &  DeepSeekMoE-3B \\
    \model{paligemma2}&  \model{google/paligemma2-3b-ft-docci-448}  & 3.03B    \\%& SigLIP-So400m/14  & Gemma2-2B \\
    \model{phi3-v}    &  \model{microsoft/Phi-3.5-vision-instruct}  &  4.15B   \\%& CLIP ViT-L/14    & Phi-3 Mini   \\
    \model{gemma3-4B} &  \model{google/gemma-3-4b-it}               & 4.3B     \\%& SigLIP-So400m ? (896x896?)   & Gemma3 (4B) \\
    \model{phi4\_mm}   &  \model{microsoft/Phi-4-multimodal-instruct} &  5.57B  \\%& SigLIP-400M & 
    \model{llava-next}&  \model{llava-hf/llava-v1.6-mistral-7b-hf}  & 7.57B    \\%& CLIP-ViT-L-336px (?)  &  Mistral-7B \\
    \model{qwen2-vl}  &  \model{Qwen/Qwen2-VL-7B-Instruct}          & 8.29B    \\%& Qwen-ViT-600M       & QwenLM \\
    \model{qwen2.5-vl}  &  \model{Qwen/Qwen2.5-VL-7B-Instruct}      & 8.29B    \\
    \model{molmo}     &   \model{allenai/Molmo-7B-O-0924}           & 7.67B    \\%& CLIP-ViT-L/14 336px &   OLMo-7B-1024  \\
    \model{idefics3}  &  \model{HuggingFaceM4/Idefics3-8B-Llama3}   & 8.46B    \\%& SigLIP-So400m/14  & Llama 3.1 \\
    \model{internvl3} &  \model{OpenGVLab/InternVL3-8B}             & 7.94B    \\%& InternViT-300M-448px-V2.5 & Qwen2.5-7B \\
    \model{gemma3-12B}&  \model{google/gemma-3-12b-it}              & 12.2B    \\%& SigLIP-So400m ? (896x896?)  & Gemma3 (12B) \\
    %\model{llama4}    &  \model{meta-llama/Llama-4-Scout-17B-16E-Instruct}     &  17B / 109B  \\ %  - & - \\
    %\model{mistral3}  &  \model{mistralai/Mistral-Small-3.1-24B-Instruct-2503} & 24B            \\ % & - & - \\
    \bottomrule
  \end{tabular}
  \caption{Models used in experiments.
  %Size is expressed in billions of parameters.
  }
  %,as found on the Hugging Face model card.}
  \label{tab:models}
\end{table*}

%   \model{deepseek\_vl\_v2} &  \model{deepseek/deepseek-vl2-tiny}              \\
%   \model{gemma3}           &  \model{google/gemma-3-4b-it}                  \\
%   \model{idefics3}         &  \model{HuggingFaceM4/Idefics3-8B-Llama3}\\
%   \model{internvl\_chat}   &  \model{OpenGVLab/InternVL2-2B}\\
%   \model{llava-next}       &  \model{llava-hf/llava-v1.6-mistral-7b-hf}\\
%   \model{paligemma2}       &  \model{google/paligemma2-3b-ft-docci-448}\\
%   \model{phi3\_v}          &  \model{microsoft/Phi-3.5-vision-instruct}\\
%   \model{qwen2\_vl}        &      \model{Qwen/Qwen2-VL-7B-Instruct} \\ 
%   % \model{qwen2\_vl}        &  \model{Qwen/Qwen-VL}\\ !!! typo in submitted version

See Table~\ref{tab:models} for the details of the models used.
Models are downloaded from HuggingFace; model details can be found at  \url{https://huggingface.co/MODEL\_NAME}.

\section{Compute}

Experiments were performed using either Nvidia A100 or A4500 GPUs.
On average, each evaluation (single model, condition) took approximately 15m, including model loading.

\section{Experimental Details}

We used the \texttt{vllm}\footnote{\url{https://docs.vllm.ai}}
package, version~\texttt{ 0.8.5.post1} with 
\texttt{transformers v4.52.0.dev0} and \texttt{torch v2.6.0}.
Models were evaluated with default settings, apart from limiting the model's output size in order to deal with memory limitations.
Generation temperature was set to $0.05$. We only evaluated the first output token.

\section{Shapley Experiments}
\label{app:shapley}

We use the ExactExplainer from the \texttt{shap}
library (version \texttt{0.48.0}): treating the image as a single part, together with the short texts,
makes this feasible. We mask the image by removing it entirely (rather than
replacing it with uniform pixels);  language tokens are masked with
\texttt{\_}.

\section{AI Agent Use}

We used coding agents (\texttt{copilot}) to assist with code development.
We did not use any AI agents for writing.

\section{Full Results}
\label{app:full-results}
Numerical results for all experiments and conditions are in Table~\ref{tab:app-full-results}.

\begin{table}
\footnotesize

\begin{tabular}{lllrrrrrr}
\toprule
prompt & image & name            & smolvlm2 & deepseek-vl-v2-tiny & paligemma2 & phi3-v & gemma3-4B  & phi3-v \\
\midrule
 concept & generic    & base      & 0.8657 & 0.7851 & 0.6427 & 0.8257 & 0.8095 & 0.7533 \\
 concept & exception  & base      & 0.4372 & 0.2755 & 0.1587 & 0.4716 & 0.3410 & 0.3929 \\
 concept & exception  & exception & 0.6833 & 0.7839 & 0.8139 & 0.6640 & 0.6971 & 0.8182 \\
 concept & generic    & exception & 0.5340 & 0.6227 & 0.7158 & 0.4785 & 0.7489 & 0.7114 \\
 instance & generic   & base      & 0.8513 & 0.6827 & 0.5778 & 0.8457 & 0.7951 & 0.7121 \\
 instance & exception & base      & 0.6540 & 0.8413 & 0.7883 & 0.6146 & 0.6833 & 0.7714 \\
 instance & exception & exception & 0.6677 & 0.8645 & 0.7664 & 0.6771 & 0.6964 & 0.8314 \\
\bottomrule
\end{tabular}

\begin{tabular}{lllrrrrr}
\toprule
 prompt & image & name            & llava-next & molmo & idefics3 & internvl3 & gemma3-12B \\
\midrule
 concept & generic    & base       & 0.9319 & 0.7858 & 0.8507 & 0.9182 & 0.8738 \\
 concept & exception  & base       & 0.6758 & 0.3829 & 0.3804 & 0.4422 & 0.3648 \\
 concept & exception  & exception  & 0.5147 & 0.8045 & 0.7164 & 0.7083 & 0.7851 \\
 concept & generic    & exception  & 0.3279 & 0.7227 & 0.4522 & 0.4934 & 0.8089 \\
 instance & generic   & base       & 0.9094 & 0.7189 & 0.8570 & 0.8932 & 0.8326 \\
 instance & exception & base       & 0.5328 & 0.8376 & 0.6964 & 0.7021 & 0.7064 \\
 instance & exception & exception  & 0.5896 & 0.8295 & 0.7002 & 0.7177 & 0.7608 \\
\bottomrule
\end{tabular}

\caption{Accuracy results for all experiment conditions.}
\label{tab:app-full-results}
\end{table}

\section{Annotation Tool}
\label{app:annotation-tool}
\begin{figure*}
\includegraphics[width=\textwidth]{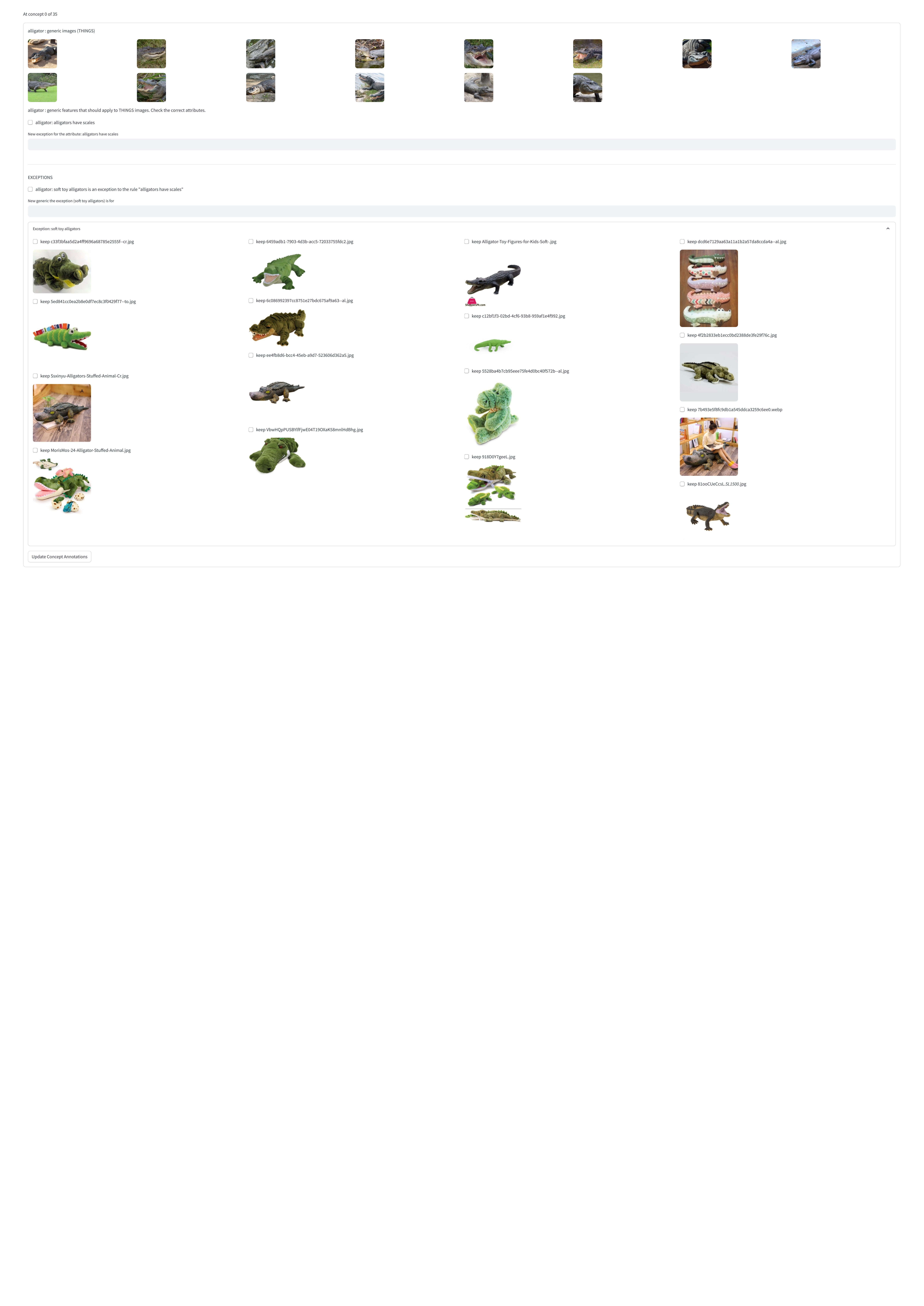}
\caption{Annotation interface for dataset validation and expansion}
\label{app:fig-interface}
\end{figure*}

See Figure~\ref{app:fig-interface}.

\end{document}